# Cementron: Machine Learning the Constituent Phases in Cement Clinker from Optical Images

Mohd Zaki[1], Siddhant Sharma[2], Sunil Kumar Gurjar[1], Raju Goyal[3,4], Jayadeva[2,5,*], N. M. Anoop Krishnan[1,5,*]

[1]Department of Civil Engineering, Indian Institute of Technology Delhi, Hauz Khas, New Delhi, India 110016

[2]Department of Electrical Engineering, Indian Institute of Technology Delhi, Hauz Khas, New Delhi, India 10016

[3]UltraTech Cement Ltd., B Wing, Ahura Centre, 2nd Floor Mahakali Caves Road, Andheri East, Mumbai, India 400093

[4]Sharda University, Plot No. 32-34, Knowledge Park III, Greater Noida, Uttar Pradesh 201310 India

[5]Yardi School of Artificial Intelligence, Indian Institute of Technology Delhi, Hauz Khas, New Delhi 110016

*Corresponding authors: krishnan@iitd.ac.in (NMAK), jayadeva@iitd.ac.in (J)

**Abstract**

Cement is the most used construction material. The performance of cement hydrate depends on the constituent phases, *viz.* alite, belite, aluminate, and ferrites present in the cement clinker, both qualitatively and quantitatively. Traditionally, clinker phases are analyzed from optical images relying on a domain expert and simple image processing techniques. However, the non-uniformity of the images, variations in the geometry and size of the phases, and variabilities in the experimental approaches and imaging methods make it challenging to obtain the phases. Here, we present a machine learning (ML) approach to detect clinker microstructure phases automatically. To this extent, we create the first annotated dataset of cement clinker by segmenting alite and belite particles. Further, we use supervised ML methods to train models for identifying alite and belite regions. Specifically, we finetune the image detection and segmentation model Detectron-2 on the cement microstructure to develop a model for detecting the cement phases, namely, Cementron. We demonstrate that Cementron, trained only on literature data, works remarkably well on new images obtained from our experiments, demonstrating its generalizability. We make Cementron available for public use.

*Keywords: Cement clinker, cement phases, optical images, detectron, computer vision*

**Introduction**

Cement, the most used construction material, is the primary binding phase of concrete [1,2]. Cement is produced in rotary kilns when raw materials like limestone, clay, and aluminosilicates are fired at temperatures beyond 1000ºC [1,2], forming clinkers, a lumpy solid mass of a few millimeters. Further, these clinkers are ground and mixed with gypsum and other additives to form cement powders. Thus, the quality of raw material, fuel used to fire the kiln, and flux added with the raw material essentially control the quality of clinker and, consequently, the performance of cement [1–8].

The main constituents of the clinker are alite, belite, aluminate, and ferrite[1,2,9–11]. The presence of alite and belite governs the strength of the cement. The presence of aluminate is detrimental to cement, as its hydration leads to the flash setting of the cement. Aluminate presence is due to the addition of aluminosilicates in clinker production. They react with the raw feed at ~1400ºC to assist in the production of silicate phases. However, ferrites are not known to cause any detrimental effect on cement and concrete [1,2]. Therefore, alite and belite are mostly desired in the clinker microstructure as they are the significant phases contributing to the concrete strength after hydration [1,2]. Traditionally, the quantification of these different phases in the clinker microscopy images is done using the point–count method [12–14].

Cement clinker microscopy became prominent due to the efforts of LeChatlier through a petrographic microscope [10,15] and followed by A. E. Tornebohm [16]. The detailed work of Ono (1995) [17] on polished clinker sections for microscopy led to the development of standard procedures for studying clinker nodules. With the advent of advanced imaging methods, researchers started using backscattered electron (BSE) images of cement microstructures [18,19] to understand their characteristics through the study of features like hydration products [20,21]. However, methods like BSE are uneconomical and time-consuming. Further, with the development of computational methods, Georget et al. (2021) recommended using machine learning (ML) for cement microstructure segmentation. Researchers also developed preprocessing steps for preparing input images for ML models [19]. Zeng et al. (2022) recently used deep learning to detect microspheres in fly ash, an important supplementary cementitious material [22].

Identifying the microstructural phases present in the cement is also challenging since the alite and belite particles have different optical characteristics [23,24]. Fig.1 shows some cement microstructure images taken from the literature. Fig.1(a) and (b) are from the same clinker [25], but the color and contrast between both images are quite different. The dark-colored crystals are alite, and the olive green-colored crystals belong to the belite class. Therefore, if a model is trained on one image to identify alite and belite crystals, it will not be able to make predictions on the other. In Fig. 1(c)[26], there are different shades of blue color within the belite crystals. Also, in brown-colored alite, there exist dark-colored lines inside the crystal. Fig.1(d)[27] shows alite and belite crystals joining together in the image's left part. In Fig. 1(e)[28], it is challenging to identify the crystal type due to the poor contrast and similar reflectance of both crystals. Finally, Fig. 1(f)[29] has alite crystals showing multiple reflectances of blue and dark brown colors. If we compare this to Fig. 1(a), where belite crystals were blue, it will be challenging for a model trained on pixel color to segment the different crystals.

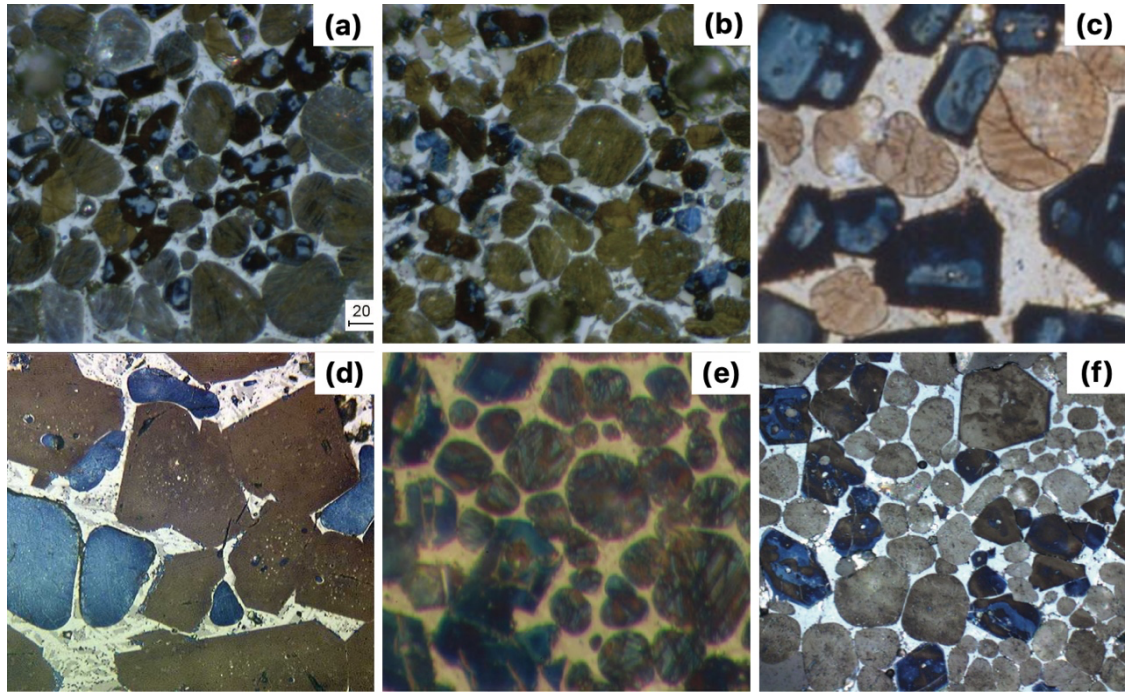

Figure 1. Challenges in cement microstructure segmentation.

ASTM C1356 standard defines the steps to perform quantitative determination of Portland cement clinker phases through the point count method [12]. After preparing the etched section, an expert must look at the section through the microscope and manually count the points [9,10]. A total of 3000 – 4000 points are required to quantify the presence of different phases in the clinker microstructure. Although some automation concerning instrumentation and software has happened, automated identification and quantification of phases from optical microscopy images of cement clinker is still challenging [30,31]. Specifically, identifying the phases of cement, that is, alite and belite remains a manual process that still relies on the skills of a domain expert. Automating this process can enable significant improvements in the quality control of cement and also accelerate physics-based computational models that require the microstructural information of cement. Although the literature is replete with methods that use ML-based segmentation to identify aggregates in construction materials [32] and the analysis of carbonate and siliciclastic thin sections [33–37], to the best of our knowledge, no application of ML for identifying alite and belite crystals in cement microstructure has been published to date. In this work, we propose two approaches for this task, namely, (a) Model on Windows (MoW), which involves training an ML model from small windows sampled from the entire microstructure image on which predictions are required, and (b) Cementron, a deep learning model based on Detectron2, an open source, state of the art model of image segmentation tasks.

**Methodology**

Fig.2 graphically shows the methodology adopted in this work. Primarily, the dataset of cement microstructure was collected from online sources. These images were then preprocessed and manually annotated to obtain the ground truth, generating the first labeled dataset on cement microstructure. These datasets were then used to train and evaluate the ML models. Detailed descriptions of the steps are given below.

***Dataset preparation:*** The cement clinker microstructure images were obtained from handbooks, research papers, and available web resources. The list of all sources from which images were taken is given in the GitHub repository of this work (https://github.com/M3RG-IITD/Cementron). To identify alite and belite particles in the microstructure images, we first performed annotation by marking their boundaries. The open-source tools: Computer Vision Annotation Tool (CVAT) [38] and labelme [39], which is a python-based image annotation tool inspired by LabelMe [40], were used for the annotation task. Further, the annotations were extracted and converted to masks for creating the dataset for training ML models. The summary of the annotated images is discussed in the results and discussion of the paper.

*Model training:*

In this work, we propose two approaches for training ML models for identifying alite and belite particles. The first approach involves training on a minimal subset of the entire image. Since this method involves taking information from small windows from the entire image, we call this approach MoW (Model on Windows). The second approach takes the input of the entire image and the corresponding mask and is trained on a larger dataset. Thus, this model is potentially generalizable to completely unseen images and can be used on new images, as demonstrated later in the results section. This model is an in-house finetuned version of Detectron2, the state-of-the-art library for computer vision-related tasks like object detection and instance segmentation [41]. This fine-tuned model for cement clinker images will now be referred to as Cementron. Note that, for both images, manual annotations are required; that is, images need to be marked with boundaries of particles to be identified. A detailed description of both models is given below.

*MoW:* In this approach, the user first captures an image of the clinker. The objective is to label each pixel in the image as alite, belite, or "other". The user first selects $m$ patches of size $n \times n$ (in terms of the number of pixels). For each pixel in the selected window, a feature vector of length $p \times p$ is created. The value of each element in this feature vector is obtained by looking up the $p \times p$ pixels surrounding the chosen central pixel. This process generates a dataset of size $m \times n^2 \times p^2$. If the image has more than one channel, say $c$ channels, the number of features in the dataset will increase from $p^2$ to $c \times p^2$. Once the dataset is created, it is split into training, validation, and test sets containing 70%, 15%, and 15%, respectively, of the data. The model is trained on the training set after tuning different hyperparameters. The model giving the best results on the validation set is finalized and evaluated on the test set.

*Cementron:* Here, the model takes the entire image as input, and extracts features from it using a feature pyramid network, also referred to as the *backbone network*. The extracted features are passed through another component of Detectron2, namely, the *region proposal network*, which provides the bounding box proposals with confidence scores corresponding to the particles detected in the image. Finally, the model generates the bounding box associated with the identified particle, the confidence score, and the segmentation mask. The backbone network was chosen as the ResNet-50 3x model in this work. Since identifying alite and belite particles was similar to the image segmentation task proposed in Microsoft Common Objects in Context (COCO) dataset, the model configuration was initialized using the resources provided in the Detectron2 GitHub repository for the image segmentation task. Further details about the model and code can be found in the GitHub repository:

https://github.com/facebookresearch/detectron2 [42].

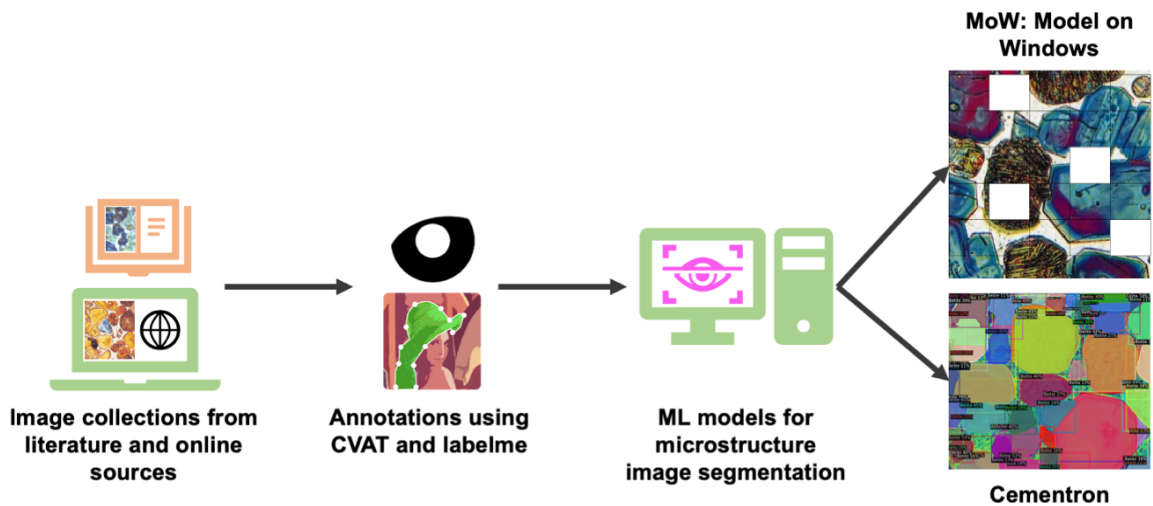

Figure 2. The methodology adopted in this work. The major steps include data collection, annotations, model training, and evaluation.

**Results and discussion**

We collected images from online sources such as handbooks and research papers to train the image segmentation models. Overall, we annotated 47 images, out of which 9 images contributed to the particles in the test set, and the remaining comprised the training set. The total number of alite particles in the training and test are 1169 and 173, respectively, and for belite, 764 and 293 particles are present in the training and test set, respectively. Thus, we annotated a total of 2,399 particles in this work. The images were selected such that those contributing to 80% of particles are in the training set, and the remaining 20% are in the test set. The training set was further subjected to 4-fold cross-validation. To ensure the quality of the ground truth dataset, the first round of annotations was cross-checked independently by two other domain experts. The corrections were made appropriately in the ground truth based on concurrence among the experts.

The annotated images were first used to train a model using the MoW approach. Note that the MoW approach is tailored for each image separately. Note that MoW is a local model; that is, we have to train separate models for each image, and each model can be used only for the image on which it is trained. In Fig. 3, we show the performance of such models on one of the microstructure images from our dataset. We train 4 different models to classify the pixels in the image as alite or belite. Here, ten windows of size $50 \times 50$ pixels are taken, which is roughly 8% of the total number of pixels in the image. Further, for each pixel in the window, we took nine features ($3 \times 3$) window, eight surrounding pixels, and one central pixel) from one channel. Since we used RGB images (3 channels), each pixel has a feature vector of length 27. Therefore, the model trained using these features will classify the pixels into different classes based on the central pixel and its neighbors. An XGBoost[43] model-based classification performed the best, and different metrics for identifying the model's capability in classifying the microstructure of these images are reported in Table 1. We observe that the model exhibits high precision and recall for both alite and belite. Note that the window should be selected so that pixels from both classes are chosen. Also, the windows are selected both in between the particles and on the edges to ensure the model can learn pixel labels from the surrounding pixels. Some places where the model makes mistakes are inside the particles that label some pixels as *others*. In the microstructure matrix, some pixels are wrongly predicted as alite or belite. It is worth noting that the MoW model can train the model well by taking only a tiny section of the original image, making this a much more straightforward approach than the manual identification of the particles.

Although the MoW approach performs reasonably well, it requires creating a dataset and training a new model each time predictions are required on a new image. Also, as seen in Fig. 1, a wide range of microstructure images are available due to variations in raw material composition, experimental approach to preparing slides for microscopy, illumination conditions, and microscopes used. A general model is required to perform well, irrespective of the constraints mentioned earlier. Hence, we train an image segmentation model based on Detectron2 for the cement microstructure images; we call this model *Cementron*, details of which are mentioned in the Methods section.

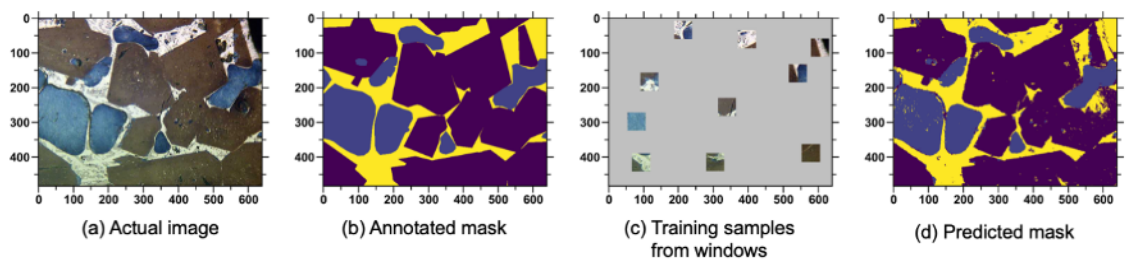

Figure 3. Visualizing steps and output for MoW approach: (a) actual image, (b) annotated mask, (c) sampling windows on the actual image, (d) predicted pixel labels

| | Precision | Recall | F1-score |
|---|---|---|---|
| Alite | 0.956 | 0.957 | 0.956 |
| Belite | 0.958 | 0.913 | 0.935 |

Table 1. XGBoost model's performance on classifying pixels as Alite and Belite for microstructure (shown in Fig.3).

We trained two sets of Cementron models separately for grayscale and RGB images. One set of models is trained to identify both alite and belite; we call it *"dual"*, and another set has two models, one each for alite and belite, called *"single"*. Thus, while the *dual* model is trained to output both alite and belite particles from the image, the *single* is trained on either alite or belite data only.

Table 2 shows the performance of the Cementron in all the cases mentioned earlier. The models trained on RGB images perform better than those trained on greyscale images. This suggests that the color of the images, instead of confusing the model due to the complexities mentioned in Fig. 1, can give additional information to make the classification better. The precision of the model trained to identify only alite particles is more than the *dual* model; however, this is not the case for belite. This trend is the opposite in recall values, where the *dual* model performed better for alite and the *single* model for belite. Looking at the F1 scores reveals that the performance of both models is quite similar. Note that Cementron also gives the confidence score for an identified particle. The scores reported in Table 2 correspond to the model at the threshold, which provided the best F1 scores.

| | Model | Precision | | Recall | | F1 Score | |
|---|---|---|---|---|---|---|---|
| | | Alite | Belite | Alite | Belite | Alite | Belite |
| Grey | single | 0.864 | 0.821 | 0.834 | 0.805 | 0.849 | 0.813 |

|  | dual | 0.937 | 0.863 | 0.79 | 0.744 | 0.857 | 0.799 |
| --- | --- | --- | --- | --- | --- | --- | --- |
| RGB | single | **0.963** | 0.861 | 0.805 | **0.876** | 0.877 | **0.868** |
|  | dual | 0.945 | **0.903** | **0.847** | 0.818 | **0.893** | 0.858 |

Table 2. Performance of Cementron models in identifying alite and belite particles from clinker microstructure images from the test dataset.

Fig.4 shows the predictions of single models trained on RGB images of different microstructures. It should be noted that the images shown in Fig. 4 are taken from the test set, i.e., these images were kept hidden from the Cementron model during the training phase. Fig. 4(a) has very poor contrast, making it difficult to segment alite and belite particles. This image was given to the trained model for identifying the pixels corresponding to alite and belite classes. The red and blue boxes on the mask correspond to the pixels wrongly labeled as alite and missed belite, respectively. The pixels mislabeled as alite are generally present at the edges. Hence, the information that alite is angular in shape is considered by the model, leading to errors. In the case of belite particles missed by the model, it is observed that these particles are very small in size, present in between alite particles or share boundary with them, and their shape is slightly angular. Despite the input image being poor in contrast, striations present inside the belite particles, and discontinuities in the alite, the Cementron model can generalize well on unseen images.

Fig. 4(b) shows the predictions using the single model trained to identify alite particles. In most cases, the model can correctly identify the pixel label as alite. In the particle in the bottom left corner, the model differentiated the speckles inside the alite and did not label it. In the particle adjacent to it, the model recognized the left part of the particle and missed the right part beyond its striation. The mistakes made by the model seem to be due to either striation in the particles or speckles.

(a)

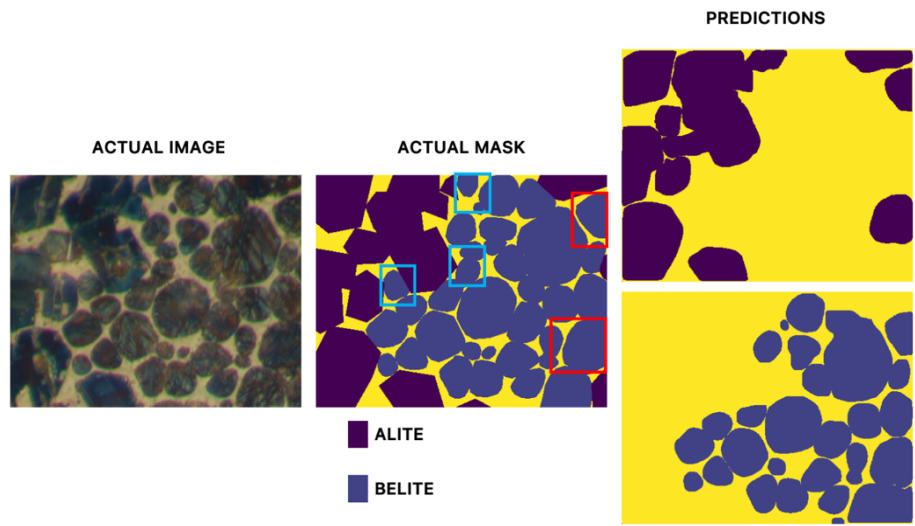

(b)

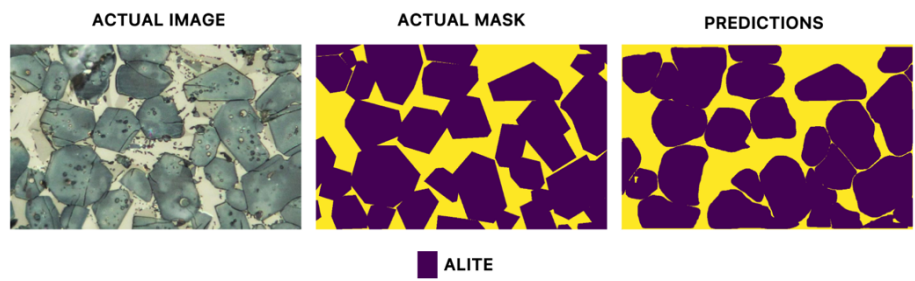

Fig. 4. Visualizing predictions of Cementron trained on RGB images for identifying individual particles: (a) image having both alite and belite, (b) image having only alite

In order to evaluate the performance of the model on an entirely new dataset, we obtained a fresh clinker sample from a cement plant in India, and the optical microscopy of these samples was performed to obtain images. Fig. 5(a) shows the optical images of this clinker, which are notably different from the images used to train the Cementron model (see Fig. 1). The blue particles are belite, and the brown ones are alite. Fig 5(c) and 5(e) show the predictions of alite and belite particles on the selected image using the respective models. The precision, recall, and F1 scores of the individual models are shown in Table 2. We observe that for a completely unseen image taken from a different source, Cementron gives excellent predictions demonstrating its ability to generalize to any optical image of the clinker. In this case, some belite and solidified melt are present as inclusions inside the alite. Due to such occurrences, the model makes a few mistakes while detecting pixels corresponding to the alite crystals, therefore, a slightly lower F1 – score. In the case of belite, the model cannot identify the belite present as inclusion in alite crystals. Also, the model missed the belite crystal of very small size, which was also observed in Fig. 4(a).

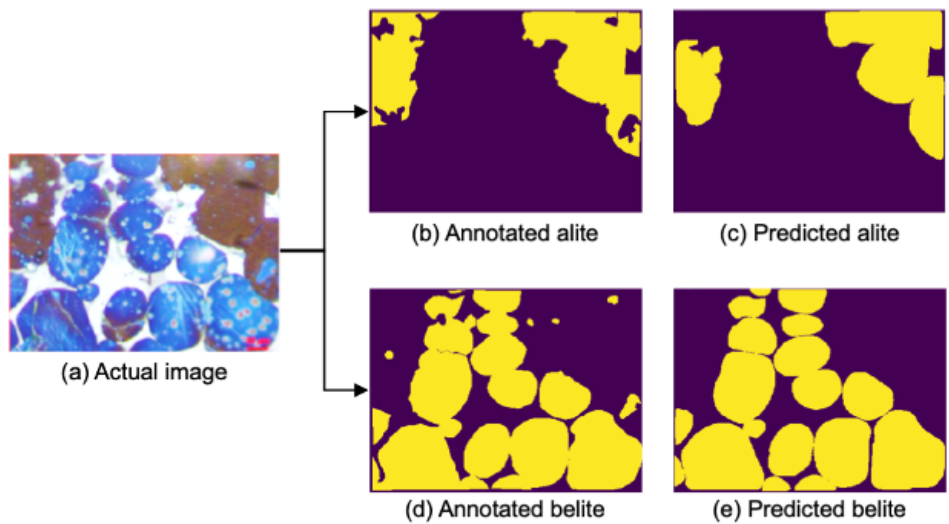

Fig. 5. Visualizing predictions and performance of Cementron model on a dataset obtained from the experimental dataset (a) actual image, (b) annotated mask for alite, (c) predicted mask for alite, (d) annotated mask for belite, (e) predicted mask for belite

|  | Precision | Recall | F1-score |
|---|---|---|---|
| Alite | 0.90 | 0.87 | 0.88 |
| Belite | 0.95 | 0.96 | 0.95 |

Table 2. Cementron's performance on classifying pixels as Alite and Belite for microstructure (shown in Fig.5).

**Applications of the developed models**

   **(a) Accelerating dataset creation**

Data is the most important and indispensable part of any ML pipeline. Manually annotating particles inside images is a tedious process. Pretrained models are generally trained on large datasets like ImageNet [44,45], whose samples are highly dissimilar to cement microstructure examples. Therefore, using such models as a backbone in the object detection and segmentation pipeline will not be able to detect alite and belite particles. Hence, the proposed model in this work, Cementron, can be used to identify the particles from cement clinker microscopy images. Although the model does misidentify some particles, errors can be corrected manually, and the pipeline saves enormous time by automatically generating annotations. The Cementron model generates masks for all particles in an image. These masks are converted into polygon masks, or run length encoding desired for preparing the dataset in the COCO format, which is required to finetune the Cementron model. The dataset and codes used in this work and the generated dataset can be accessed at the GitHub repository for this work: https://github.com/M3RG-IITD/Cementron

   **(b) Particle size analysis and accelerated point counting**

Since Cementron predicts labels for individual particles along with the pixel mask, it is possible to get the number of particles and the area covered by each particle type if the scale of the image is known. In Fig. 6(a), we identify the center of the detected alite particles in the clinker microstructure (Fig. 4), label them with an asterisk, and provide the particle ID. To obtain particle size distribution curves for the microstructure shown in Fig. 6 (a), we calculate the area and size of a given particle by counting the number of pixels in it and the length of the diagonal of the bounding box enclosing the particle. To compare the particle distribution obtained from both cases, we normalized both sets of values to the same scale, ranging from 0.1 to 16. The y-axis shows the percentage of particles finer than the normalized particle size shown on the x-axis. It can be seen from the curve that both metrics can capture the particle size distribution of the given microstructure. Note that the particle size distribution of clinkers is an important quantity as their reactivity is directly proportional to their particle sizes. In addition, the particle size distribution can also potentially provide insights into the process parameters during the production of the clinker.

(a)

(b)

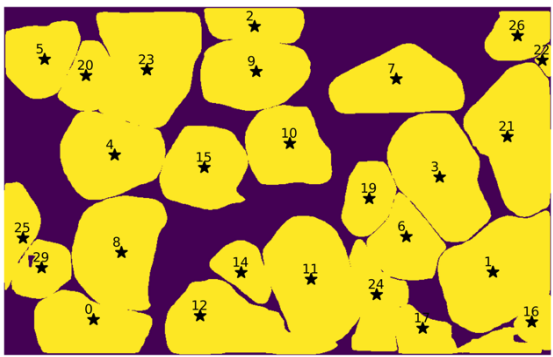
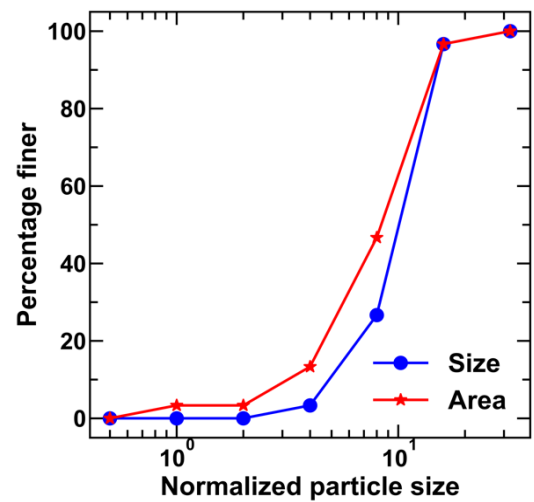

Fig. 6. Clinker microstructure with identified particles (a) centers are marked with a '*' symbol, and the number alongside the symbol indicates the assigned particle's id, (b) particle size distribution curve obtained using particle size and area.

### (c) Generating numerical model of cement clinker

The strength of the cement paste depends upon the hydration of different phases present in the clinker. To simulate the hydration process, many tools and strategies are proposed in the literature, such as "*CEMHYD3D*", "*µic,*" and "*HYDCEM*" [46–49]. With the help of Cementron and postprocessing procedures, the clinker microstructure can be converted into the numerical model as shown in Fig. 7. Here, the masked image of the clinker microstructure is meshed using the tool Nanomesh [50], which converts the digital images into meshes for use as input for simulations. From the generated microstructure, nodes are inserted at the edges of detected particles. These nodes are used by the conforming Delaunay triangulation algorithm [51] implemented in Nanomesh [50] to generate the mesh as shown in Fig. 7. This microstructure can, in turn, be used to generate realistic microstructure that can, in turn, be used for thermodynamic simulations of the clinker hydration process.

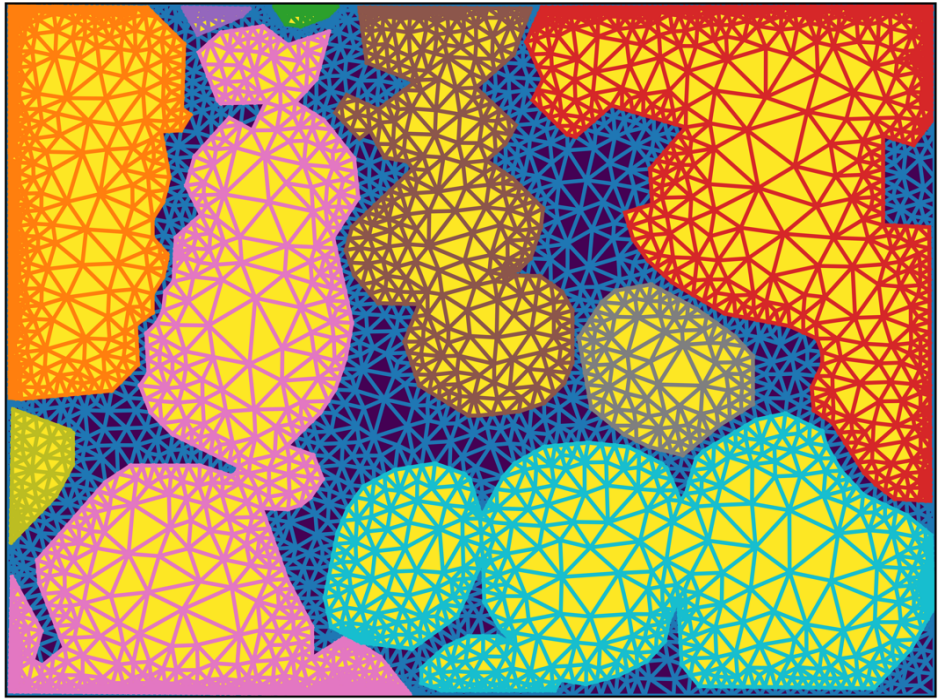

Fig. 7. An example of creating numerical microstructure of cement clinker. The orange and red colored meshes on the upper left and right side of the image represent the alite particles, and the blue-colored mesh between the particles represents the matrix. All other meshed particles are belite.

**Conclusion**

This work presents two approaches for segmenting cement clinker microstructure images. First, (MoW), requires annotation and training for each new image. The second, Cementron, requires a large dataset of annotated images but results in a general model and yields good predictions on unseen images. Creating an annotated dataset for the Cementron model is difficult, but this is a one-time effort that results in developing a generalizable model and can save researchers time in the long run. Identifying alite and belite crystals from microstructures is also important for cement industries in monitoring and controlling cement quality. Knowing the number of different phases present in the clinker allows different parameters of clinker production, like temperatures, raw feed, and cooling rates, to be adjusted to produce the right quality cement. A model like Cementron can help create a numerical model from the microstructure images, which can be used in simulations to model properties like strength, hydration, and thermal effects, to name a few. It is a well-known fact that creating annotated datasets for image segmentation tasks is time-consuming and requires human supervision. Therefore, the annotated dataset used in this work has been shared on the GitHub repository as a first step in developing larger manually annotated cement databases. Note that although the work is currently applied to clinker microstructure, this can be extended to cement and concrete systems to analyze the microstructure quantitatively and develop realistic models that can be used in numerical simulations.

**References:**

[1] H.F. Taylor, Cement chemistry, Thomas Telford London, 1997.
[2] P. Hewlett, M. Liska, Lea's chemistry of cement and concrete, Butterworth-Heinemann, 2019.
[3] K.L. Scrivener, Options for the future of cement, Indian Concr. J. 88 (2014) 11–21.
[4] S. Krishnan, S. Bishnoi, High level clinker replacement in ternary limestone-calcined clay-clinker cement, in: Advances in Structural Engineering, Springer, 2015: pp. 1725–1731.
[5] S. Ghosh, Advances in cement technology: critical reviews and case studies on manufacturing, quality control, optimization and use, Elsevier, 2014.
[6] K.L. Scrivener, V.M. John, E.M. Gartner, Eco-efficient cements: Potential economically viable solutions for a low-CO2 cement-based materials industry, Cement and Concrete Research. 114 (2018) 2–26. https://doi.org/10.1016/j.cemconres.2018.03.015.
[7] G. Habert, S.A. Miller, V.M. John, J.L. Provis, A. Favier, A. Horvath, K.L. Scrivener, Environmental impacts and decarbonization strategies in the cement and concrete industries, Nat Rev Earth Environ. 1 (2020) 559–573. https://doi.org/10.1038/s43017-020-0093-3.

[8] A.C. Emmanuel, S. Bishnoi, Effect of curing temperature and clinker content on hydration and strength development of calcined clay blends, Advances in Cement Research. (2022) 1–14.

[9] F. Hofmänner, Microstructure of Portland Cement Clinker. Holderbank Management and Consulting, Ltd., Holderbank/CH, Holderbank Management and Consulting, Ltd., Holderbank/CH. (1975).

[10] D.H. Campbell, Microscopical examination and interpretation of Portland cement and clinker, Portland Cement Assn, 1986.

[11] J. Bhatty, F. Miller, S. Kosmatka, Innovations in portland cement manufacturing'Portland cement association, Skokie, IL. (2004).

[12] C01 Committee, Test Method for Quantitative Determination of Phases in Portland Cement Clinker by Microscopical Point-Count Procedure, ASTM International, n.d. https://doi.org/10.1520/C1356-07R20.

[13] M.J. Neilson, G.F. Brockman, The error associated with point-counting, American Mineralogist. 62 (1977) 1238–1244.

[14] R.J. Howarth, Improved estimators of uncertainty in proportions, point-counting, and pass-fail test results, American Journal of Science. 298 (1998) 594–607.

[15] R.H. Bogue, The chemistry of Portland cement, LWW, 1955.

[16] P. Stutzman, Scanning electron microscopy imaging of hydraulic cement microstructure, Cement and Concrete Composites. 26 (2004) 957–966. https://doi.org/10.1016/j.cemconcomp.2004.02.043.

[17] Y. Ono, Ono's method: Fundamental microscopy of portland cement clinker, Chichibu Onoda Cement Corporation, 1995.

[18] K.L. Scrivener, P.L. Pratt, Characterisation of Portland Cement Hydration by Electron Optical Techniques, MRS Online Proceedings Library (OPL). 31 (1983) 351. https://doi.org/10.1557/PROC-31-351.

[19] F. Georget, W. Wilson, K.L. Scrivener, edxia: Microstructure characterisation from quantified SEM-EDS hypermaps, Cement and Concrete Research. 141 (2021) 106327. https://doi.org/10.1016/j.cemconres.2020.106327.

[20] K.L. Scrivener, Backscattered electron imaging of cementitious microstructures: understanding and quantification, Cement and Concrete Composites. 26 (2004) 935–945. https://doi.org/10.1016/j.cemconcomp.2004.02.029.

[21] M. Mouret, E. Ringot, A. Bascoul, Image analysis: a tool for the characterisation of hydration of cement in concrete – metrological aspects of magnification on measurement, Cement and Concrete Composites. 23 (2001) 201–206. https://doi.org/10.1016/S0958-9465(00)00061-5.

[22] Z. Zeng, Y. Wei, Z. Wei, W. Yao, C. Wang, B. Huang, M. Gong, J. Yang, Deep learning enabled particle analysis for quality assurance of construction materials, Automation in Construction. 140 (2022) 104374. https://doi.org/10.1016/j.autcon.2022.104374.

[23] P.E. Stutzman, Microscopy of Clinker and Hydraulic Cements, Reviews in Mineralogy and Geochemistry. 74 (2012) 101–146. https://doi.org/10.2138/rmg.2012.74.3.

[24] I. Maki, S. Ito, T. Tanioka, Y. Ohno, K. Fukuda, Clinker grindability and textures of alite and belite, Cement and Concrete Research. 23 (1993) 1078–1084. https://doi.org/10.1016/0008-8846(93)90167-8.

[25] W. Dai, C. Gong, L. Lu, X. Cheng, EFFECT OF MgO ON CALCINATION AND PROPERTIES OF BELITE-BARIUM CALCIUM SULPHOALUMINATE CEMENT CLINKER WITH $Na_2O$ AND $K_2O$, Ceramics - Silikaty. (2018) 121–130. https://doi.org/10.13168/cs.2018.0003.

[26] New clinkers, International Cement Review. (2014). https://www.cemnet.com/Articles/story/153986/new-clinkers.html (accessed November 6, 2022).

[27] RM88A50.jpg (640×484), (n.d.). http://publish.illinois.edu/concretemicroscopylibrary/files/2014/05/RM88A50.jpg (accessed November 6, 2022).

[28] P.E. Tsakiridis, G.D. Papadimitriou, S. Tsivilis, C. Koroneos, Utilization of steel slag for Portland cement clinker production, Journal of Hazardous Materials. 152 (2008) 805–811. https://doi.org/10.1016/j.jhazmat.2007.07.093.

[29] B. Hökfors, Phase chemistry in process models for cement clinker and lime production (Doctoral thesis), (2014).

[30] PETROG Digital Petrography, (n.d.). https://ws2.petrog.com/index.html (accessed September 27, 2022).

[31] Automatic Point Counter with Moving Slide Holder, PELCON. (n.d.). http://pelcon.dk/products/point-counter/ (accessed September 27, 2022).

[32] H. Huang, J. Luo, E. Tutumluer, J.M. Hart, A.J. Stolba, Automated Segmentation and Morphological Analyses of Stockpile Aggregate Images using Deep Convolutional Neural Networks, Transportation Research Record. 2674 (2020) 285–298. https://doi.org/10.1177/0361198120943887.

[33] A. Koeshidayatullah, M. Morsilli, D.J. Lehrmann, K. Al-Ramadan, J.L. Payne, Fully automated carbonate petrography using deep convolutional neural networks, Marine and Petroleum Geology. 122 (2020) 104687. https://doi.org/10.1016/j.marpetgeo.2020.104687.

[34] S. Pattnaik, S. Chen, A. Helba, S. Ma, Automatic carbonate rock facies identification with deep learning, 2020. https://doi.org/10.2118/201673-MS.

[35] X. Liu, H. Song, Automatic identification of fossils and abiotic grains during carbonate microfacies analysis using deep convolutional neural networks, Sedimentary Geology. 410 (2020) 105790. https://doi.org/10.1016/j.sedgeo.2020.105790.

[36] R.A. Rubo, C. de Carvalho Carneiro, M.F. Michelon, R. dos S. Gioria, Digital petrography: Mineralogy and porosity identification using machine learning algorithms in petrographic thin section images, Journal of Petroleum Science and Engineering. 183 (2019) 106382. https://doi.org/10.1016/j.petrol.2019.106382.

[37] A. Buono, S. Fullmer, K. Luck, K. Peterson, H. king, P. More, S. LeBlanc, Quantitative digital petrography: Full thin section quantification of pore space and grains, 2019. https://doi.org/10.2118/194899-MS.

[38] Computer Vision Annotation Tool (CVAT), (2021). https://github.com/openvinotoolkit/cvat (accessed December 16, 2021).

[39] K. Wada, Labelme: Image Polygonal Annotation with Python, (2022). https://doi.org/10.5281/zenodo.5711226.

[40] LabelMe annotation tool source code, (2022). https://github.com/CSAILVision/LabelMeAnnotationTool (accessed September 27, 2022).

[41] facebookresearch/detectron2, (2022). https://github.com/facebookresearch/detectron2 (accessed September 27, 2022).

[42] Welcome to detectron2's documentation! — detectron2 0.6 documentation, (n.d.). https://detectron2.readthedocs.io/en/latest/ (accessed September 27, 2022).

[43] T. Chen, C. Guestrin, XGBoost: A Scalable Tree Boosting System, Proceedings of the 22nd ACM SIGKDD International Conference on Knowledge Discovery and Data Mining. (2016) 785–794. https://doi.org/10.1145/2939672.2939785.

[44]   J. Deng, W. Dong, R. Socher, L.-J. Li, K. Li, L. Fei-Fei, ImageNet: A large-scale hierarchical image database, in: 2009 IEEE Conference on Computer Vision and Pattern Recognition, 2009: pp. 248–255. https://doi.org/10.1109/CVPR.2009.5206848.

[45]   A. Krizhevsky, I. Sutskever, G.E. Hinton, ImageNet classification with deep convolutional neural networks, Commun. ACM. 60 (2017) 84–90. https://doi.org/10.1145/3065386.

[46]   D.P. Bentz, D.P. Bentz, CEMHYD3D: A three-dimensional cement hydration and microstructure development modelling package. Version 2.0, US Department of Commerce, National Institute of Standards and Technology, 2000.

[47]   S. Bishnoi, K.L. Scrivener, μic: A new platform for modelling the hydration of cements, Cement and Concrete Research. 39 (2009) 266–274. https://doi.org/10.1016/j.cemconres.2008.12.002.

[48]   N. Holmes, D. Kelliher, M. Tyrer, Simulating cement hydration using HYDCEM, Construction and Building Materials. 239 (2020) 117811. https://doi.org/10.1016/j.conbuildmat.2019.117811.

[49]   Y. Cao, M. Kazembeyki, L. Tang, N.M.A. Krishnan, M.M. Smedskjaer, C.G. Hoover, M. Bauchy, Modeling the nanoindentation response of silicate glasses by peridynamic simulations, J. Am. Ceram. Soc. 104 (2021) 3531–3544. https://doi.org/10.1111/jace.17720.

[50]   Renaud, Nicolas, Smeets, Stef, Corbijn van Willenswaard, Lars J., Nanomesh: A Python workflow tool for generating meshes from image data, (2022). https://doi.org/10.5281/ZENODO.7157382.

[51]   J.R. Shewchuk, Triangle: Engineering a 2D quality mesh generator and delaunay triangulator, in: Workshop on Applied Computational Geometry, 1996: pp. 203–222.